\documentclass[11pt]{article}

\usepackage[]{acl}
\usepackage{times}
\usepackage{latexsym}
\usepackage{enumerate}
\usepackage[T1]{fontenc}

\usepackage[utf8]{inputenc}

\usepackage{microtype}

\usepackage{inconsolata}

\usepackage{graphicx}

\usepackage{amsmath}
\usepackage{listings}
\usepackage{multirow}
\usepackage{booktabs}
\usepackage{algorithm}
\usepackage{algorithmic} 
\usepackage{needspace}

\title{Knowledge-driven Augmentation and Retrieval for Integrative Temporal Adaptation}

\author{Weisi Liu\\
  University of Memphis \\
  \texttt{wliu9@memphis.edu} \\\And
  Guangzeng Han \\
  University of Memphis \\
  \texttt{ghan@memphis.edu} \\\And
  Xiaolei Huang \\
  University of Memphis \\
  \texttt{xiaolei.huang@memphis.edu} \\
  }

\begin{document}
\maketitle
\begin{abstract}
Time introduces fundamental challenges in model development and deployment: models are usually trained on historical data while deployed on future data where semantic distributions and domain knowledge may evolve.
Unfortunately, existing studies either overlook temporal shifts or hardly capture rich shifting patterns of both semantic and knowledge.
We develop Knowledge-driven Augmentation and Retrieval for Integrative Temporal Adaptation (KARITA) to capture diverse temporal shifts (e.g., uncertainty and feature shift), construct and integrate rich knowledge sources (e.g., medical ontology like MeSH), and leverage shifting insights for selecting-retrieval augmented learning. 
We evaluate KARITA on classification tasks across multiple domains, clinical, legal, and scientific corpora, demonstrating consistent improvements across multiple domains with temporal adaptation. 
Our results show that knowledge integration can be more critical and effective in temporal augmentation and learning.
\end{abstract}



\begin{table*}[thp]
\centering
\resizebox{1\textwidth}{!}{
\begin{tabular}{l|l|l|l|l}
\multicolumn{1}{c}{\textbf{Dataset}} 
& \multicolumn{1}{c}{\textbf{Time Span}} 
& \multicolumn{1}{c}{\textbf{Task}} 
& \multicolumn{1}{c}{\textbf{Labels}} 
& \multicolumn{1}{c}{\textbf{Data Size}} \\ \hline \hline

MIMIC    
& 2008--2019      
& Phenotype inference             
& Top 50 frequent ICD-10 codes    
& 331,794 clinical notes \\

EurLex   
& 1958--2016      
& Legal topic classification      
& 21 Top-level EuroVoc categories 
& 65,000 legal documents \\

arXiv-CS 
& 1991--2025      
& Scientific paper categorization 
& Top 30 CS subject categories    
& 622,419 paper abstracts \\

\end{tabular}
}
\caption{Overview of three time-varying datasets with time span, task, labels, and data size.}
\label{tab:dataset}
\end{table*}

\section{Introduction}

Standard training and validation settings commonly overlook varying patterns between training and real-world deployment by randomly splitting data into multiple sets with the same distribution. 
Time plays a critical role in evolving data and its distributions when data corpora span over years and seasons, where we can call the varying patterns between training and test as \textit{temporal shifts}. 
Studies have shown such temporal shifts can be complex and its cause can be heterogeneous~\cite{liu2025time,agarwal2022temporal}.
For example, ~\cite{liu2025time} found that biomedical classifiers perform much worse when time spans are closer while other tasks may perform worse when time spans are farther~\cite{liu2025examining}; and such patterns may vary across other domains.  
Therefore, a question to be answered is: \textit{How can we capture the complexity and heterogeneity of temporal shift and further mitigate its effects on models?}

\textit{Temporal learning}, a framework to understand and integrate temporal shifts, aims to promote consistent model performance over periods of time.
In contrast to test-time adaptation focusing on discrete data splits~\cite{shi2024medadapter}, the framework will learn  evolving patterns over multiple continuous time splits.
Existing studies in this track primarily follow changing semantic meanings or representations, such as word sense shifting~\cite{su2022improving} or feature embedding distances~\cite{huang2019neural,rottger2021temporal}.
For example, \citet{huang2019neural} model temporal shifts by learning diachronic word embeddings and adapting classifiers across time intervals, treating changes in word usage and representation spaces as the primary signal of temporal variation.
However, the unified feature representations may not be adequate to fully capture the diversity and heterogeneity of temporal shifts and therefore mislead model learning steps, particularly when corpus domains are versatile and high-stakes.
For example, in clinical and legal document classification, temporal shifts may stem from multiple co-occurring factors, such as evolving practices, changing conventions, and shifts in disease prevalence, with different types of shifts dominating at different time periods. Treating these heterogeneous shifts as a single unified change can obscure their individual impacts and lead to unstable model performance over time.
Thus, a second question to be answered by our study is: \textit{can temporal learning better inform and keep model robust over time?}
To answer the questions, we present \textit{K}nowledge-driven \textit{A}ugmentation and \textit{R}etrieval method for \textit{I}ntegrative \textit{T}emporal \textit{A}daptation (\textit{KARITA}) that discover shifting patterns from diverse aspects, enable time-aware learning process, and promote data quality by knowledge-driven augmentations over time. 
We introduce three major modules to achieve our goals, multi-aspect shift detector, source backtracking retrieval, and data augmentation.
We construct domain knowledge and validate our approach with multiple state-of-the-art methods in temporal learning ~\cite{santosh2024chronoslex}, closer work ~\cite{agarwal2022temporal}, and without adapting time ~\cite{niu2022EATA,niu2023towards_SAR} over health, legal, and scientific domains.
Our results highlight temporal shifts come diverse sources beyond regular embedding features, inform model with temporal shifts can be critical, and domain knowledge is efficient to augment model performance in temporal learning.\footnote{Our code is available at ~\url{https://github.com/trust-nlp/TemporalLearning-KARITA}.}

% ...


\section{Related Work}

\paragraph{Data Drift}



Data drift is common in real-world NLP applications, and prior studies have repeatedly observed that temporal mismatches between training and test data can impair model performance~\cite{agarwal2022temporal,liu2025examining}
Such drift can stem from multiple sources: e.g., word senses shifting over time~\cite{kutuzov2018diachronic,tang-etal-2023-word}, changes in data distributions~\cite{guo2023predict}, or factual knowledge updates as real-world entities and relationships evolve~\cite{jang2022towards}.
Correspondingly, many studies have focused on specific types of shifts to improve model generalization, including topic shift~\cite{huang2018modeling}, lexical semantic adaptation~\cite{su2022improving}, temporal concept shift~\cite{chalkidis2022improved,margatina2023dynamic}, and overall data distribution drift~\cite{guo2023predict}.
However, these different types of shifts do not typically occur in isolation; rather, they often co-occur across time periods.
Methods that focus on a single shift type may overlook other concurrent changes, thereby limiting adaptation effectiveness.
In contrast, our method explicitly detects multiple shift patterns and addresses them through targeted data augmentation  that combines the generative capabilities of large language models~\cite{han2025attributes,rao2026scoping} and external domain Knowledge.





\paragraph{Temporal Learning}


Existing efforts to address temporal drift can be broadly categorized into three directions. 
\emph{Model-centric} approaches modify network architectures to explicitly encode time or capture time-varying patterns~\citep{liu2025examining}. 
\emph{Feature-centric} methods focus on learning temporally robust or aligned representations to mitigate distributional mismatch across time~\citep{dhingra2022time}. 
\emph{Training-centric} solutions redesign learning strategies, including time-aware training curricula~\citep{santosh2024chronoslex} and continual learning frameworks~\citep{rottger2021temporal, agarwal2022temporal, shang2022improving}. 
In contrast, \emph{data-centric} approaches that explicitly manipulate training data to cope with temporal drift remain relatively underexplored as a primary lever for temporal learning.



A common limitation of most existing approaches is that temporal adaptation is often treated as a one-time or stage-wise adjustment process, implicitly assuming relatively clear boundaries between temporal regimes~\citep{liu2025examining,rottger2021temporal}. 
However, in realistic scenarios, distributional drift is typically gradual, overlapping, and heterogeneous, which challenges such assumptions and limits the effectiveness of one-time adaptation strategies.
Meanwhile, retrieval-based methods have shown strong potential in adapting models to evolving information needs across tasks~\cite{wang2025reasoningretrieval,ji2025mrag,ji2026retrieval}.
We propose to adapt models to temporal shift through backtracking retrieval from historical data and knowledge-driven augmentation.
Our approach departs from prior work in two key aspects. 
First, instead of performing one-off adaptation, we model temporal learning as an iterative and selective process that continuously identifies and reacts to emerging shifts.
Second, we adopt a data-centric perspective that explicitly retrieves and augments data conditioned on detected temporal shifts.

\section{Data}

Biomedical, legal, and scientific domains are characterized by continuous evolution of terminology and knowledge, which poses substantial challenges for text classification models deployed over time. 
We select three corpora from these domains to systematically study temporal shifts and adaptation in multi-label classification tasks: phenotype inference from clinical notes (MIMIC-IV-Notes), legal topic classification (EurLex), and scientific document classification (arXiv-CS).
Table~\ref{tab:dataset} reports the main statistics of the datasets. 

\textbf{MIMIC-IV-Notes} (MIMIC)~\cite{johnson2023mimic-note, goldberger2000physiobank} is a collection of de-identified clinical notes including discharge summaries and radiology reports. 
We choose the discharge summaries and focus on the phenotype inference task.
Each discharge summary is annotated with International Classification of Diseases (ICD) codes, which indicate the presence of diseases, symptoms, injuries, and other health conditions.
The earlier notes are annotated with ICD-9 codes, and we convert them to ICD-10 using a standard ICD mapping toolkit~\footnote{https://github.com/snovaisg/ICD-Mappings} for consistency.
We choose the 50 most frequent ICD-10 codes as labels. 
The de-identification procedure in MIMIC-IV assigns each note to a three-year time interval, spanning from 2008 to 2022, naturally forming five temporal domains. 
Due to data sparsity in the final interval, We only use the first four intervals (2008–2019) in our experiments. 
Detailed data partitioning is provided in Appendix~\ref{sec:temporal_partition}.

\textbf{EurLex}~\cite{chalkidis2021multiEurLex} is a large-scale legal corpus consisting of European Union laws published between 1958 and 2016. 
The documents are annotated with concepts from the EuroVoc~\cite{walhain-etal-2025-eurovoc} taxonomy by the Publications Office of the European Union. 
Each EuroVoc concept is associated with a semantic descriptor, and documents are originally labeled with one or more concepts drawn from different levels of the hierarchy.
EuroVoc is organized into eight hierarchical levels. 
In this work, we focus on the top level of the EuroVoc concept, which contains 21 legal domain categories (e.g. Economics), and focus the English portion of the corpus. 

\textbf{arXiv-CS} is a scientific document corpus derived from arXiv abstracts in the computer science category, covering publications from 1991 to 2025. 
Each paper is associated with one or more subject categories that reflect its research area (e.g., \textit{cs.LG}, \textit{cs.CV}, \textit{cs.AI}). 
We choose the 30 most frequent computer science subdomains as the labels.



\begin{figure*}[htp]
\centering
\includegraphics[width=0.99 \textwidth]{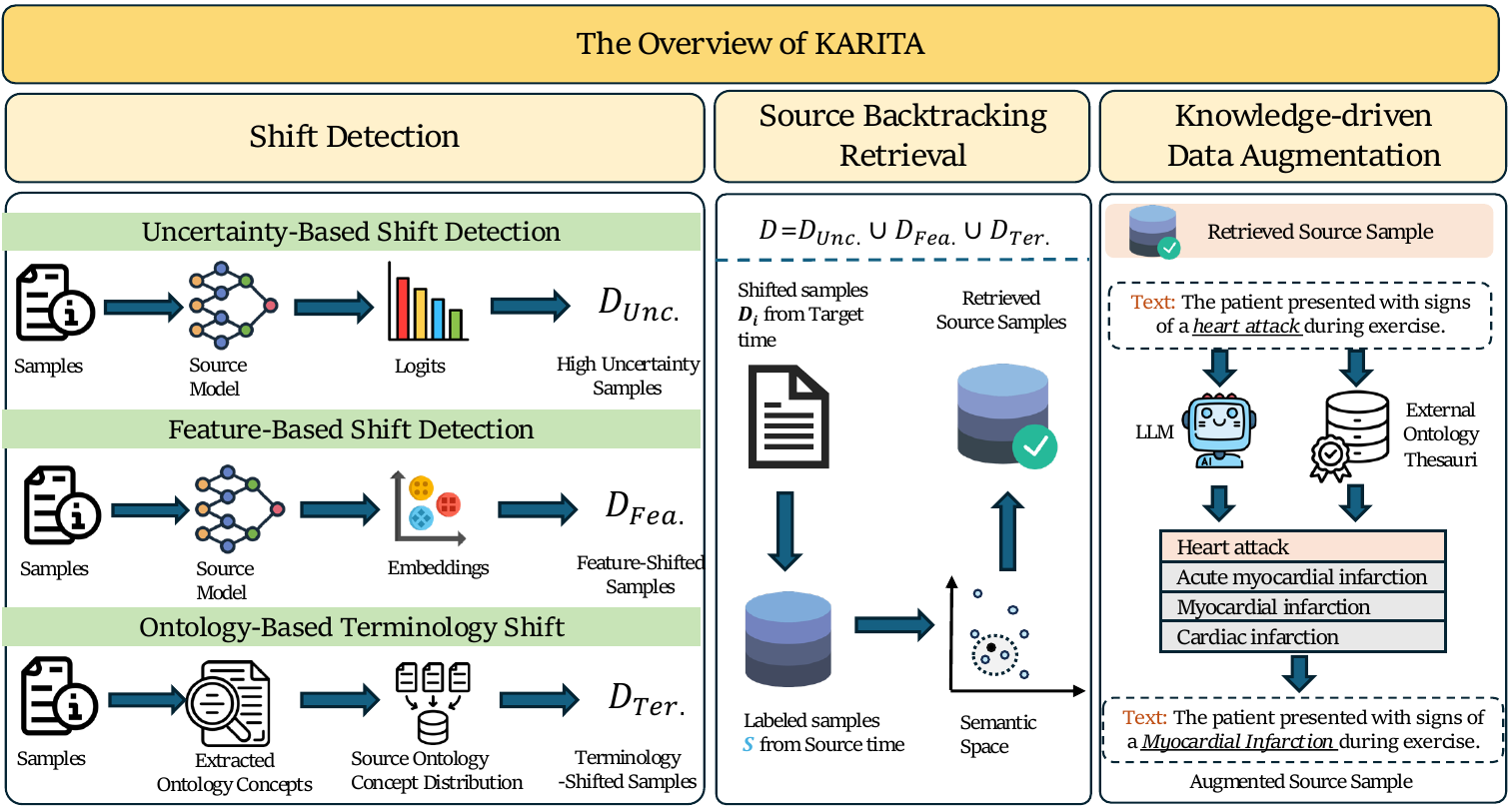} 
\caption{The KARITA method overview. For each incoming target batch, the model performs shift detection, retrieves semantically related source instances, and applies knowledge-driven data augmentation.
The augmented data are then used to update the model parameters, and the same process is repeated for the next target batch.
}
\label{fig:karita}
\end{figure*}




\section{Method}

In this section,  we present our KARITA framework in Figure~\ref{fig:karita}.
The key idea is to automatically detect shifts from target data and use the detected shifted samples to retrieve similar source samples and use LLM augmented knowledge and external ontology thesaurus to augment the retrieved source data, and use the augmented data to adapt the model.
It consists of three major modules: 1) Shift Detection, 2) Source Backtracking Retrieval, and 3) Knowledge-driven Data Augmentation.
We include detailed workflow in Algorithm~\ref{alg:karita} under the Appendix.

\subsection{Shift Detection}
For a target instance $x$, we compute three normalized shift scores
capturing uncertainty, feature-level deviation, and ontology-based terminology drift. 
These scores are then combined to determine whether
$x$ requires retrieval.

\paragraph{Uncertainty-Based Shift}
Model uncertainty is often associated with distribution shift, since out-of-distribution samples are typically harder to classify.
Maximum predicted probability reflects the model's confidence, while predictive entropy measures prediction dispersion.
Combining them provides a reliable indicator of inputs likely to deviate from the training distribution.
Let $p_l(x) = \sigma(z_l(x))$ be the sigmoid probability for label $l$,
where $z_l(x)$ is the model's logit output.
We define the maximum predicted probability as
$p(x) = \max_l \, p_l(x)$
and the average binary entropy as
$
H(x) = \frac{1}{L}\sum_{l=1}^{L} -\big[p_l(x)\log p_l(x) + (1-p_l(x))\log(1-p_l(x))\big],
$
where $L$ is the total number of labels.
We define a binary uncertainty indicator:
$
U(x) = \mathbf{1}\big[p(x) < \tau_p \;\wedge\; H(x) > \tau_H \big].
$
We set $\tau_p=0.5$ (i.e.\ empty prediction threshold) and $\tau_H=0.25$ in all experiments.

\paragraph{Feature-Based Shift}

Data shift can manifests in the representation space, where changes in input distributions lead to deviations in learned feature embeddings.
Measuring feature-level deviation therefore provides a meaningful proxy for detecting distribution mismatch.
Mahalanobis distance~\cite{mahalanobis1936generalised} has been shown effective for out-of-distribution detection~\cite{Lee2018Advances}, and we adopt it here to measure feature-level deviation from the source embedding statistics.
Let $E(x)$ denote the embedding of $x$ and 
$\mu, \Sigma$ be the mean and covariance of source-domain sample embeddings.
The Mahalanobis distance is
$
d(x) = \sqrt{(E(x)-\mu)^\top \Sigma^{-1}(E(x)-\mu)}.
$
We normalize it into $[0,1]$ via
$
F(x) = \mathrm{clip}\!\left(
\frac{d(x) - d_{\min}}{d_{\max} - d_{\min} + \varepsilon},\; 0,\; 1
\right),
$
where $d_{\min}, d_{\max}$ are computed from source-domain sample embeddings.

\paragraph{Ontology-Based Terminology Shift}
Ontology shift reflects how domain-specific terminology changes over time, where newly emerging concepts or shifts in usage frequency push test samples away from the ontology distribution observed during training, thereby increasing the risk of model degradation when such terms were unseen or underrepresented in early data. 
To quantify this shift, we treat the source ontology as a probability distribution over concepts and measure how strongly a target document deviates from it. 
Let $\mathcal{C}(x)$ denote the set of ontology concepts detected in document $x$, and define $p_{t_1}(c)=f_{t_1}(c)/N_{t_1}$ as the source-period document-frequency probability of concept $c$, where $f_{t_1}(c)$ is its count and $N_{t_1}$ is the total number of source documents. Inspired by information theoretic surprisal, we compute $I(c)=-\log(p_{t_1}(c)+\varepsilon)$, where a larger value indicates rarer or unseen concepts relative to the training distribution, hence more likely to cause temporal shift. We then aggregate across all concepts in a target document to obtain the ontology-tail shift score
$$
O_{\text{tail}}(x)=\frac{1}{|\mathcal{C}(x)|}\sum_{c\in\mathcal{C}(x)}-\log(p_{t_1}(c)+\varepsilon),
$$
where a greater $O_{\text{tail}}(x)$ indicates that the document relies on long-tail terminology under the source ontology distribution. 
This formulation gives a continuous notion of terminology drift rather than relying on the maximum over concepts, and naturally highlights documents whose vocabularies fall in the distribution tail, making them strong candidates for targeted augmentation and test-time adaptation.

\paragraph{Detection Trigger}
We use the uncertainty trigger to directly collect target samples 
satisfying $U(x)=1$, while feature- and ontology-based metrics 
produce continuous scores and are used to rank samples. 
For $F(x)$ and $O(x)$, we select the top $\rho$ fraction 
(e.g., $\rho=10\%$) of high-scoring target samples:
$
\mathcal{D}_{F} = \text{Top}_{\rho}(F(x)),
\mathcal{D}_{O} = \text{Top}_{\rho}(O(x)).
$
Let $\mathcal{D}_{U}=\{x\mid U(x)=1\}$ be the set activated by 
uncertainty. The final detected shift set is
$
\mathcal{D}_{shift} = \mathcal{D}_{U} \cup \mathcal{D}_{F} \cup \mathcal{D}_{O},
$
which is then passed to the retrieval module for augmentation.


We set $\rho=0.1$ (top 10\%) in all experiments
(see Appendix~\ref{sec:sensitivity} for sensitivity analysis).

\subsection{Source Backtracking Retrieval}


Temporal shifts do not necessarily imply the complete absence of relevant information in historical data. 
Even when domain-specific terminology or feature distributions evolve over time, target instances may still share underlying semantic patterns with source data.
As a result, shifted target samples often admit semantically similar counterparts from the source period.
This motivates us to propose source backtracking retrieval:
this module identifies semantically similar source samples for each detected shifted target instance.
The retrieved samples provide semantically relevant source-domain information used for model adaptation under temporal shift.


Specifically, given a target sample $x_t$, we first obtain its semantic representation (e.g. the \texttt{[CLS]} token
) $\mathbf{z}_t = f_{\Theta_s}^{enc}(x_t)$, $\Theta_s$ is the source-trained model.
We compute source sample embedding $\mathbf{z}_s$ using the same encoder.
We then measure the semantic similarity between target and source samples using cosine similarity in the embedding space:
$
\text{sim}(x_t, x_s) = \cos(\mathbf{z}_t, \mathbf{z}_s).
$
For each target instance, we retrieve the top-$k$ source samples with the highest similarity scores. In this paper, $k$ is set to three (see Appendix~\ref{sec:sensitivity} for sensitivity analysis).
These retrieved samples serve as semantically aligned source references that remain relevant under temporal shift, and are subsequently used for knowledge-driven data augmentation in the next stage.

\begin{table*}[htp]
\centering
\resizebox{1\textwidth}{!}{
\begin{tabular}{l|lllll|lllll|lllll}
Dataset       & \multicolumn{5}{c|}{MIMIC-IV-Notes}    & \multicolumn{5}{c|}{EurLex}            & \multicolumn{5}{c}{arXiv-CS}          \\
\hline
Method        & P     & R     & sa-F1 & mi-F1 & ma-F1 & P     & R     & sa-F1 & mi-F1 & ma-F1 & P     & R     & sa-F1 & mi-F1 & ma-F1 \\
\hline \hline
\multicolumn{16}{c}{I. Source Pretrained Performance on Source} \\ 
Source Model    & 66.83 & 51.92 & 55.40 & 59.31 & 47.52 & 86.09 & 80.39 & 80.92 & 81.11 & 66.97 & 66.69 & 58.46 & 60.56 & 62.36 & 43.65 \\ \hline
\multicolumn{16}{c}{I. Target Pretrained Performance on Target}  \\ 
Target Model    & 73.63 & 57.91 & 62.12 & 76.66 & 65.78 & 89.27 & 85.28 & 85.73 & 86.10 & 71.74 & 81.60 & 76.05 & 74.98 & 72.18 & 65.51 \\ \hline
\multicolumn{16}{c}{II. Source Pretrained Performance on Target}   \\ 
Source Model    & 58.78 & 44.07 & 47.41 & 52.86 & 40.65 & 79.65 & 56.86 & 63.73 & 64.48 & 46.75 & 52.12 & 40.21 & 43.36 & 45.85 & 34.86 \\ \hline
\multicolumn{16}{c}{III. Source Model Performance on Target after adaptation}                                                                        \\
Ours          & \textbf{64.14} & \textbf{62.13} & \textbf{60.15} & \textbf{63.95} & \textbf{52.12} & \textbf{82.79} & \textbf{69.49} & \textbf{73.09} & \textbf{73.71} & \textbf{56.15} & \textbf{68.14} & \textbf{64.83} & \textbf{62.63} & \textbf{61.55} & \textbf{49.82} \\
IFT           & 58.60 & 41.56 & 49.64 & 55.24 & 43.05 & 81.17 & \underline{52.08} & \underline{60.54} & \underline{61.92} & \underline{37.12} & 58.43 & 46.23 & 49.17 & 49.75 & 40.67 \\
SAR           & 59.41 & 40.97 & 45.68 & 51.96 & 36.65 & \underline{73.71} & 60.47 & 63.84 & 64.21 & 48.30 & 53.80 & 40.20 & 43.40 & 50.60 & 38.10 \\
Self-Labeling & 58.52 & 43.52 & 46.99 & 52.34 & 40.55 & 80.84 & 54.22 & 61.62 & 62.42 & 42.02 & 52.25 & 40.33 & 43.46 & 45.95 & 34.94 \\
EATA          & \underline{55.85} & \underline{34.75} & \underline{40.37} & \underline{45.98} & \underline{28.02} & 77.40 & 58.92 & 64.32 & 64.83 & 47.97 & \underline{41.40} & \underline{33.29} & \underline{34.90} & \underline{37.84} & \underline{27.63}
\end{tabular}
}
\caption{Overall classification performance (\%) on MIMIC-IV-Notes, EurLex, and arXiv-CS before and after adaptation. 
All results are averaged over 3 runs using different random seeds. Results compare base models fine-tuned on source data (source models) evaluated on both source and target test sets, base models fine-tuned on target data (target models) as upper bounds, and source models adapted to the target domain using KARITA and baseline adaptation methods. We \textbf{bold} the highest and \underline{underline} the lowest performance of each column.}
\label{tab:performance}
\end{table*}

\subsection{Knowledge-driven Data Augmentation}



In many classification settings, label assignment depends on knowledge-bearing expressions whose surface forms may evolve over time.
Pretrained language models, however, may have limited exposure to such evolving expressions, leading to degraded performance when terminology or phrasing shifts across time.
To address this, we apply knowledge-driven data augmentation to the retrieved source samples associated with shifted target instances by aligning the terminology between the source domain and target domain.
From a representation perspective, synonym-based augmentation can be viewed as a form of controlled lexical perturbation.  
This encourages invariance to terminology variation and improves robustness to temporal shift.

We employ two complementary sources of synonym knowledge for data augmentation: 
1) LLM-based task-relevant Term identification and synonym generation and 2) external ontology thesauri.
For the EurLex and arXiv-CS datasets, we prompt LLM to identify terminology relevant to label assignment and generate corresponding synonyms.
For the MIMIC-IV-Notes dataset, due to privacy constraints on non-open-source models, we rely exclusively on external ontology thesauri to provide structured terminological relations.

\paragraph{Task-Relevant Term Identification and Synonym Mapping}
For EurLex and arXiv-CS documents, we employ GPT-4o-mini\footnote{https://openai.com/index/gpt-4o-mini-advancing-cost-efficient-intelligence/} to analyze target texts and identify task relevant terms that are informative for the classification task.
Based on these terms, the LLM provides corresponding synonymous or historically used variants, and this forms a mapping between current expressions and alternative surface forms.
Each target instance is provided together with the set of candidate labels to guide term identification.
We provide target instance along with the set of all candidate labels to GPT-4o-mini.
The prompt template used for experiments is detailed in Appendix~\ref{sec:prompt}.


\paragraph{External Ontology Thesauri}
To ensure comprehensive and reliable terminological coverage, we incorporate domain-specific ontology thesauri that provide structured synonym relationships: Medical Subject Headings (MeSH), EuroVoc and Computer Science Ontology (CSO).
Details on the specific versions and processing procedures for these resources are provided in Appendix~\ref{sec:external_resources}.

\textbf{MeSH}~\cite{mesh2025} is the controlled and hierarchically-organized vocabulary thesaurus produced by the U.S. National Library of Medicine. It's widely used for indexing, cataloging, and searching for biomedical and health-related information.

\textbf{EuroVoc}~\cite{walhain-etal-2025-eurovoc} is a multidisciplinary thesaurus maintained by the Publications Office of the European Union. It contains keywords organized into 21 domains and 127 sub-domains, used to describe and index the content of legal and policy documents for the European Union.
 
\textbf{CSO}~\cite{salatino2020cso} is a large-scale ontology of Computer Science research areas, comprising approximately 14,000 topics and 162,000 semantic relationships, automatically generated from scholarly publications in Computer Science.


\section{Experiment}
To examine the effectiveness of our proposed KARITA framework under temporal shift, we conduct experiments on three specialized multi-label classification datasets across different time periods.
We compare KARITA with a source model trained on historical data, several state-of-the-art baselines, and an target model fine-tuned directly on target-time data as an upper bound.

All methods are initialized from the same source model trained on the earliest time interval and are evaluated on target time-domain test sets.
Performance is evaluated using both instance-level and label-level multi-label classification metrics. Specifically, we report sample-averaged precision, recall, and F1 score to assess prediction quality at the sample level, along with micro- and macro-averaged F1 scores to measure overall performance across labels under varying degrees of label imbalance. 
All results are reported on the target time test sets.
This experimental design allows us to assess (i) the extent of performance degradation caused by temporal shift and (ii) how effectively different adaptation strategies mitigate this degradation.










\subsection{Baselines}
We compare our method with representative baselines that cover three major paradigms for temporal adaptation: training-based adaptation, data-centric pseudo-labeling, and test-time adaptation (TTA).

\textbf{IFT}~\cite{santosh2024chronoslex} is a training-based temporal adaptation method that updates the model sequentially across time. 
This strategy improves upon traditional fine-tuning by exposing the model to chronologically ordered training data incrementally, while keeping the model architecture and loss function unchanged.


\textbf{Self-Labeling}~\cite{agarwal2022temporal} represents a data-centric adaptation approach. 
It first applies the source-trained model to generate pseudo-labels for target-domain samples. These silver-labeled target instances are then combined with the original gold-labeled source data to further fine-tune the model. 

\textbf{SAR}~\cite{niu2023towards_SAR} and \textbf{EATA}~\cite{niu2022EATA} are test-time adaptation (TTA) methods that update the model online during inference without access to target labels. 
Both methods are built upon the TENT~\cite{wang2021tent} framework, which adapts models by minimizing prediction entropy at test time. 
SAR improves robustness by incorporating sharpness-aware optimization to stabilize model updates under distribution shift, while EATA selectively updates the model using only reliable low-entropy samples and constrains parameter drift to alleviate catastrophic forgetting.

\begin{table*}[h]
\centering
\resizebox{1\textwidth}{!}{
\begin{tabular}{l|lllll|lllll|lllll}
Dataset       & \multicolumn{5}{c|}{MIMIC-IV-Notes}    & \multicolumn{5}{c|}{EurLex}            & \multicolumn{5}{c}{arXiv-CS}          \\ \hline
Method          & P     & R     & sa-F1 & mi-F1 & ma-F1 & P     & R     & sa-F1 & mi-F1 & ma-F1 & P     & R     & sa-F1 & mi-F1 & ma-F1 \\ \hline \hline
Full Method             & 64.14 & \textbf{62.13} & \textbf{60.15} & \textbf{63.95} & \textbf{52.12} & \textbf{82.79} & \textbf{69.49} & \textbf{73.09} & \textbf{73.71} & \textbf{56.15} & 68.14 & \textbf{64.83} & \textbf{62.63} & \textbf{61.55} & \textbf{49.82} \\
w/o detection    & 64.67 & 57.98 & 58.18 & 61.88 & 49.33 & 79.60 & 69.03 & 71.39 & 71.69 & 48.77 & 65.36 & 54.80 & 56.56 & 56.62 & \underline{31.02} \\
w/o augmentation & \textbf{66.75} & \underline{57.53} & 58.85 & \underline{61.80} & \underline{48.13} & 80.33 & 67.55 & 70.78 & 71.43 & 54.60 & \textbf{68.79} & 62.59 & 62.01 & 61.01 & 43.74 \\
w/o retrieval     & \underline{62.77} & 59.49 & \underline{58.08} & 62.00 & 50.67 & \underline{75.57} & \underline{65.83} & \underline{67.30} & \underline{68.88} & \underline{44.16} & \underline{57.68} & \underline{54.09} & \underline{52.40} & \underline{52.89} & 36.40
\end{tabular}
}
\caption{Ablation study on the contribution of each module. We \textbf{bold} the highest and \underline{underline} the lowest performance in each column.}
\label{tab:ablation}
\end{table*}

\begin{figure*}[htp]
\centering
\includegraphics[width=\textwidth]{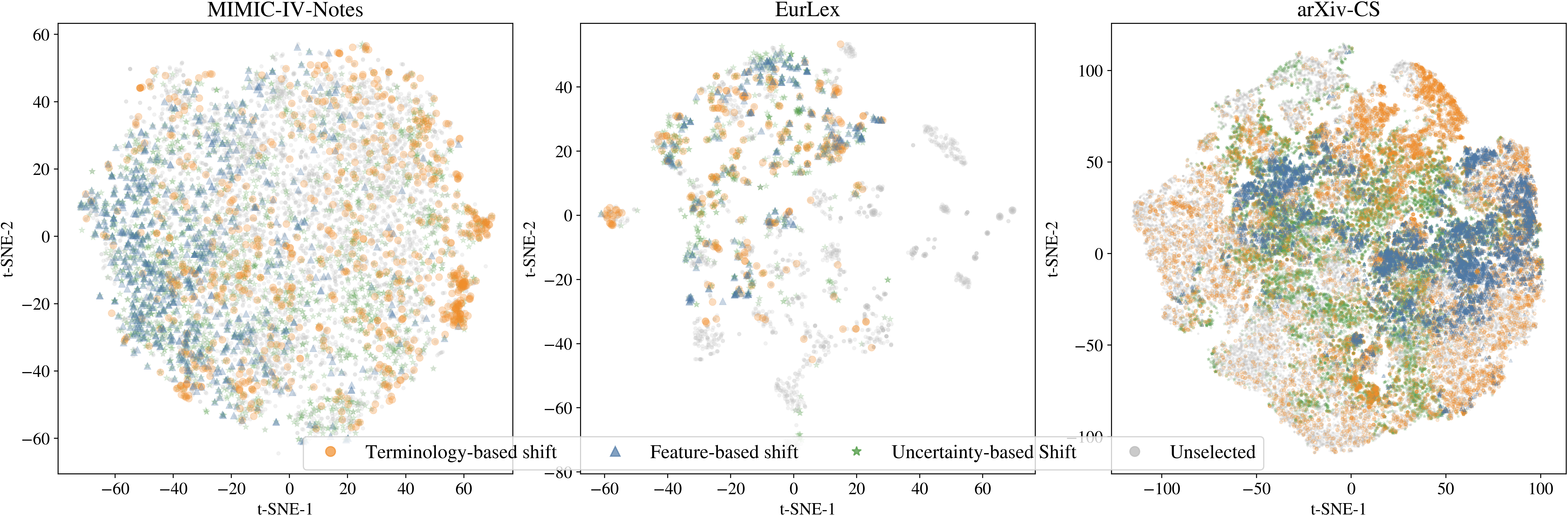} 
\caption{t-SNE visualization of target-domain representations on MIMIC-IV-Notes, EurLex, and arXiv-CS. Samples detected by ontology-based shift (orange circles), feature-based shift (blue triangles), and uncertainty-based shift (green stars) are highlighted, with all other target samples shown in grey.}
\label{fig:tsne}
\end{figure*}

\section{Results}


As shown in Table \ref{tab:performance}, our proposed method consistently achieves the state-of-the-art (SOTA) performance across all datasets and metrics in the adaptation scenarios. Specifically, compared to the source model evaluated directly on the target domain (Section II), our method yields substantial improvements, e.g., an absolute increase of \textbf{11.47\%} in ma-F1 on MIMIC-IV-Notes and \textbf{14.96\%} on arXiv-CS. This demonstrates the high effectiveness of our adaptation strategy in bridging the domain gap.

The consistent gains across diverse domains suggest that our strategy of shift-aware retrieval followed by synonym-based augmentation effectively addresses the limitations of purely unsupervised or pseudo-label-based adaptation. By retrieving semantically similar samples from the source domain, the model leverages reliable ground-truth labels to bridge the distributional gap, thereby avoiding the error accumulation common in self-labeling or the instability inherent in entropy-minimization-based test-time adaptation (e.g., the performance degradation of EATA on MIMIC-IV-Notes). This targeted alignment allows the model to capture target-specific features while maintaining the high-quality supervision signals preserved from the source domain, resulting in particularly robust performance in recall-sensitive tasks.
\subsection{Ablation Study}
We conduct an ablation study to verify the necessity of each component in our framework. As shown in Table~\ref{tab:ablation}, removing any single module leads to a consistent performance drop across all three datasets. 
Specifically, replacing our shift-aware retrieval with the purposeful selection of low-similarity samples (\textit{w/o retrieval}) results in the \textbf{lowest sa-F1 scores} across all datasets (e.g., 58.08\% on MIMIC-IV-Notes and 52.40\% on arXiv-CS). This suggests that without semantically relevant "bridges" from the source domain, the model struggles to maintain prediction quality at the individual sample level. It highlights that semantic relevance is not just beneficial for overall distribution alignment, but fundamental for effective per-sample knowledge transfer.

Furthermore, replacing our detection module with random sampling (\textit{w/o detection}) significantly impairs the model's ability to adapt, with the macro-F1 on arXiv-CS plummeting from 49.82\% to 31.02\%. This suggests that identifying specific shift dimensions is critical for selecting the "correct" knowledge to transfer. 
Removing augmentation (\textit{w/o augmentation}) also yields inferior results compared to the full method. This is most evident on arXiv-CS, where ma-F1 drops from 49.82\% to 43.74\%, suggesting that terminology alignment contributes beyond what retrieval alone provides. Overall, the synergy between these three modules is what drives robust adaptation capabilities.


\section{Analysis}


In this section, we examine how the three shift signals in KARITA relate to each other and contribute to adaptation. 
We first investigate their spatial distribution in the feature space, asking whether they capture redundant or complementary information. 
We then quantify their pairwise overlap on target samples and  track how each type of shift evolves over time. 
Finally, we compare adaptation performance when each shift detector is used in isolation versus in combination.

\subsection{How do different types of data shift relate in the feature space?}

Figure~\ref{fig:tsne} shows t-SNE projections of target-domain representations, with samples detected by each shift type marked in different colors.
We observe that ontology-based shift consistently highlights regions in the target-domain embedding space that are weakly captured by feature-based and uncertainty-based shift signals.
Across all three datasets, ontology-shifted samples form distinct spatial patterns, including clusters that are separated from feature-shifted regions, suggesting that terminology-level changes are not always reflected as representation drift.

This effect is most evident in MIMIC, where feature-based and ontology-based shifts concentrate along different directions of the embedding space. While feature shift follows a broad global trend, ontology shift forms several localized clusters with limited overlap. This suggests that relying solely on feature-based shift may miss samples whose difficulty arises from evolving medical terminology, as such changes do not necessarily produce detectable deviations in the feature space.

A similar separation is observed in EurLex, where ontology-shifted samples form compact clusters that are only weakly aligned with feature-based shift, despite the overall sparsity of the embedding.
This suggests that legal terminology evolution can introduce temporal challenges that are not fully captured by representation-level deviation.

In arXiv-CS, feature-based and uncertainty-based shifts mainly concentrate in dense regions of the embedding space, whereas ontology-shifted samples span a much broader area and blend more closely with samples considered unshifted.
This pattern indicates that terminology-driven label ambiguity can persist even when documents remain close in feature space, further motivating the need for ontology-aware signals.
Overall, these observations show that ontology-based shift is not redundant with feature- or uncertainty-based criteria, but reveals terminology-driven temporal effects that would otherwise remain difficult to detect.

\subsection{How do the three shift signals complement each other?}

To further quantify the relationship among the three shift
detectors, we examine both their overlap on detected target samples
and the temporal trends of their respective metrics on the target time period.

\begin{table}[h]
\centering
\resizebox{\columnwidth}{!}{
\begin{tabular}{l|c|c|ccc|c}
Dataset & $|\mathcal{D}_t|$ & U
        & U$\cap$O & U$\cap$F & O$\cap$F
        & U$\cap$O$\cap$F \\
\hline \hline
MIMIC    & 6,827   & 27.74\%
         & 3.05\% & 6.09\% & 0.37\% & 0.13\% \\
EurLex   & 2,542   & 27.50\%
         & 4.48\% & 8.03\% & 1.49\% & 1.26\% \\
arXiv-CS & 113,495 & 25.77\%
         & 2.71\% & 4.38\% & 1.45\% & 0.62\% \\
\end{tabular}
}
\caption{Overlap among shift detectors, reported as the percentage of target-domain samples ($|\mathcal{D}_t|$) detected by each signal or their intersections. 
U denotes uncertainty-based shift; O and F denote
ontology-based and feature-based shift, both detecting
the top $\rho{=}10\%$ of $\mathcal{D}_t$.}
\label{tab:overlap}
\end{table}

\paragraph{Overlap across shift types}
Table~\ref{tab:overlap} reports the sample counts and
pairwise intersections among uncertainty-based (U),
ontology-based (O), and feature-based (F) shift detections.
Across all three datasets, the pairwise intersections are
small: on MIMIC, only 3.05\% of target samples fall under
both U and O, and the three-way intersection (U$\cap$O$\cap$F)
accounts for merely 0.13\%.
The overlap between ontology shift and the other two types is
particularly limited (e.g., O$\cap$F = 0.37\% on MIMIC), confirming that terminology-level drift captures a distinct
aspect of temporal shift that is missed by
feature- or uncertainty-based signals.


\begin{table}[h]
\centering
\resizebox{\columnwidth}{!}{
\begin{tabular}{l|cc|cc|cc}
 & \multicolumn{2}{c|}{F score} & \multicolumn{2}{c|}{O score} & \multicolumn{2}{c}{Entropy} \\
Year & mean & med. & mean & med. & mean & med. \\
\hline \hline
\multicolumn{7}{c}{\textit{EurLex (2011--2015)}} \\ \hline
2011 & .551 & .585 & 1.536 & 1.493 & .194 & .177 \\
2012 & .579 & .629 & 1.561 & 1.489 & .209 & .209 \\
2013 & .576 & .599 & 1.596 & 1.560 & .206 & .199 \\
2014 & .601 & .643 & 1.585 & 1.548 & .213 & .213 \\
2015 & .580 & .637 & 1.526 & 1.491 & .207 & .206 \\ \hline
\multicolumn{7}{c}{\textit{arXiv-CS (2021--2025)}} \\ \hline
2021 & .655 & .686 & 1.357 & 1.350 & .134 & .137 \\
2022 & .659 & .690 & 1.358 & 1.355 & .135 & .138 \\
2023 & .667 & .696 & 1.367 & 1.362 & .136 & .139 \\
2024 & .670 & .696 & 1.391 & 1.388 & .136 & .138 \\
2025 & .673 & .698 & 1.427 & 1.424 & .138 & .140 \\
\end{tabular}
}
\caption{Year-wise statistics of feature shift (F score), ontology shift (O score), and predictive entropy in the target period for EurLex and arXiv-CS.}
\label{tab:yearwise}
\end{table}

\begin{table*}[h]
\centering
\resizebox{1\textwidth}{!}{
\begin{tabular}{l|lllll|lllll|lllll}
Dataset       & \multicolumn{5}{c|}{MIMIC-IV-Notes}    & \multicolumn{5}{c|}{EurLex}            & \multicolumn{5}{c}{arXiv-CS}          \\ \hline
Method & P  & R  & sa-F1  & mi-F1  & ma-F1  & P & R & sa-F1 & mi-F1 & ma-F1 & P & R & sa-F1 & mi-F1 & ma-F1 \\ \hline \hline
Full Method                  & 64.14 & \textbf{62.13} & \textbf{60.15} & \textbf{63.95} & \textbf{52.12} & \textbf{82.79} & \textbf{69.49} & \textbf{73.09} & \textbf{73.71} & \textbf{56.15} & \textbf{68.14} & \textbf{64.83} & \textbf{62.63} & \textbf{61.55} & \textbf{49.82} \\
Feature-based Shift Only     & \underline{63.73} & 59.23 & 58.61 & 63.84 & 51.58 & 77.33 & \underline{57.54} & \underline{62.96} & \underline{63.4}  & \underline{44.57} & \underline{54.28} & \underline{51.43} & \underline{49.59} & \underline{52.01} & \underline{23.00}    \\
Ontology-based  Shift Only   & 65.42 & 56.24 & 57.26 & 59.99 & \underline{40.94} & \underline{72.45} & 69.32 & 68.32 & 68.98 & 50.48 & 59.92 & 54.43 & 53.78 & 55.51 & 29.69 \\
Uncertainty-Based Shift Only & \textbf{67.96} & \underline{50.72} & \underline{55.15} & \underline{58.36} & 42.45 & 82.11 & 68.65 & 71.78 & 72.27 & 54.97 & 67.68 & 56.24 & 58.25 & 57.03 & 42.64 \\ 
\end{tabular}
}
\caption{Performance of using individual shift detectors versus combining all three. We \textbf{bold} the highest and \underline{underline} the lowest performance of each column.}
\label{tab:shift_types}
\end{table*}

\paragraph{Temporal trends of shift scores}
To examine whether the three signals capture temporal
evolution consistently, we compute year-wise statistics of
feature shift scores (F score), ontology shift scores
(O score), and predictive entropy for EurLex and arXiv-CS target time data (Table~\ref{tab:yearwise}).
We exclude MIMIC because its
de-identification procedure maps timestamps to
three-year intervals and year-level breakdowns is
unavailable.

Across both datasets, all three scores trend upward over
time, indicating that temporal distance from the source
period amplifies all three types of shift.
Notably, the scores increase at different rates: on EurLex,
ontology scores rise more steeply than feature scores,
while on arXiv-CS the two grow at comparable rates.
This suggests that while all three detectors respond to
increasing temporal distance, they do so with different
sensitivities depending on the domain, further motivating
their joint use.


\subsection{How do different types of data shift affect adaptation performance?}

Table~\ref{tab:shift_types} compares adaptation performance when only a single type of shift is detected. 
Feature-based shift detection performs the worst on both EurLex and arXiv-CS, with a large drop in macro-F1. For example, on arXiv-CS, macro-F1 declines to 23.00\%, compared to 49.82\% achieved by the full method. 
This shows that relying solely on feature or embedding distance fails to identify shifts that are helpful for adaptation and can even harm performance for certain label categories.

Uncertainty-based shift detection performs the worst on MIMIC-IV-Notes, where macro-F1 drops to 42.45\%, well below the full method (52.12\%). 
This shows that outcome-driven shift detection based on predictive uncertainty (e.g., high entropy or low maximum probability) is useful but not sufficient on its own, and combining it with text-level shift detection leads to more effective adaptation.

By contrast, ontology-based shift detection leads to more stable performance across datasets. As shown in Fig.~\ref{fig:tsne}, ontology-aware signals capture semantically meaningful shifts that are not clearly separated in the feature space and provide more useful samples for retrieval and augmentation.
Overall, these results show that combining complementary shift signals is necessary for robust adaptation.


%
\section{Conclusion}
We introduced \textbf{KARITA}, a knowledge-driven augmentation  framework for integrative temporal adaptation. KARITA detects temporal shift signals from multiple complementary perspectives (uncertainty, feature deviation, and ontology-based terminology drift), backtracks to retrieve semantically aligned labeled instances from historical source data, and performs knowledge-driven synonym augmentation using external thesauri and LLM-based terminology mapping. Across clinical, legal, and scientific corpora, KARITA consistently improves target-time performance over strong temporal adaptation baselines, highlighting that temporally robust learning benefits from both shift-aware sample selection and domain knowledge integration.

Our analyses show that different shift types occupy distinct regions in the representation space and contribute differently to the adaptation process, indicating that a single drift signal is often insufficient for reliable temporal learning. By unifying multi-aspect shift detection and terminology-level augmentation, the proposed KARITA framework provides a flexible, data-centric approach that is model-agnostic and effective across domains, offering a promising direction for maintaining deployed language models under evolving language.

\section*{Limitations}

While KARITA demonstrates effective temporal adaptation across classification tasks in three major domains, two important limitations should be acknowledged:
First, the proposed framework relies on the availability of external terminological resources, such as large language models or curated ontology thesauri, to construct mappings between alternative expressions.
In highly specialized, low-resource, or privacy-sensitive domains where well-developed ontologies or non-open-source LLMs are unavailable, this requirement may reduce direct applicability.
In such cases, alternative solutions, including domain-adapted language models, expert-curated terminological mappings, or corpus-induced synonym discovery from the data itself, may be necessary to maintain the effectiveness.

Second, KARITA primarily addresses temporal shifts manifested at the lexical and terminological level.
While this captures a common and impactful form of temporal variation, the framework does not explicitly model deeper forms of knowledge evolution, such as the emergence of entirely new concepts, changes in label definitions, or shifts in task formulation over time.
Extending the approach to handle such structural knowledge changes remains an important direction for future work.

\section*{Acknowledgment}
The authors thank anonymous reviewers for their insightful feedback. 
The project was partially supported by the National Science Foundation (NSF) under awards TI-2434589 (OpenAI API expenses) and IIS-2440381.
We thank the computing resources provided by the iTiger GPU cluster~\cite{sharif2025ITIGER} supported by the NSF MRI program under the award CNS-2318210.

\bibliography{ref}

@inproceedings{Lee2018Advances,
 author = {Lee, Kimin and Lee, Kibok and Lee, Honglak and Shin, Jinwoo},
 booktitle = {Advances in Neural Information Processing Systems},
 editor = {S. Bengio and H. Wallach and H. Larochelle and K. Grauman and N. Cesa-Bianchi and R. Garnett},
 pages = {},
 publisher = {Curran Associates, Inc.},
 title = {A Simple Unified Framework for Detecting Out-of-Distribution Samples and Adversarial Attacks},
 url = {https://proceedings.neurips.cc/paper_files/paper/2018/file/abdeb6f575ac5c6676b747bca8d09cc2-Paper.pdf},
 volume = {31},
 year = {2018}
}

@inproceedings{mahalanobis1936generalised,
  title={On the generalised distance in statistics},
  author={MAHALANOBIS, PC},
  booktitle={Proceedings of the National Institute of Science of India},
  volume={12},
  pages={49--55},
  year={1936}
}

@misc{sharif2025ITIGER,
      title={Cultivating Multidisciplinary Research and Education on GPU Infrastructure for Mid-South Institutions at the University of Memphis: Practice and Challenge}, 
      author={Mayira Sharif and Guangzeng Han and Weisi Liu and Xiaolei Huang},
      year={2025},
      eprint={2504.14786},
      archivePrefix={arXiv},
      primaryClass={cs.DC},
}

@misc{mesh2025,
  author = {{U.S. National Library of Medicine}},
  title = {Medical Subject Headings ({MeSH})},
  year = {2025},
  howpublished = {\url{https://www.nlm.nih.gov/mesh/}},
  note = {Accessed: 2025-01-03}
}

@inproceedings{salatino2019classifier,
author = {Salatino, Angelo A. and Osborne, Francesco and Thanapalasingam, Thiviyan and Motta, Enrico},
title = {The CSO Classifier: Ontology-Driven Detection of Research Topics in Scholarly Articles},
year = {2019},
isbn = {978-3-030-30759-2},
publisher = {Springer-Verlag},
address = {Berlin, Heidelberg},
url = {https://doi.org/10.1007/978-3-030-30760-8_26},
doi = {10.1007/978-3-030-30760-8_26},
booktitle = {Digital Libraries for Open Knowledge: 23rd International Conference on Theory and Practice of Digital Libraries, TPDL 2019, Oslo, Norway, September 9-12, 2019, Proceedings},
pages = {296–311},
numpages = {16},
location = {Oslo, Norway}
}

@InProceedings{chalkidis2021multieurlex,
  author = {Chalkidis, Ilias  
                and Fergadiotis, Manos
                and Androutsopoulos, Ion},
  title = {MultiEURLEX -- A multi-lingual and multi-label legal document 
               classification dataset for zero-shot cross-lingual transfer},
  booktitle = {Proceedings of the 2021 Conference on Empirical Methods
               in Natural Language Processing},
  year = {2021},
  publisher = {Association for Computational Linguistics},
  location = {Punta Cana, Dominican Republic},
  url = {https://arxiv.org/abs/2109.00904}
}

@inproceedings{walhain-etal-2025-eurovoc,
    title = "The {E}uro{V}oc Thesaurus: Management, Applications, and Future Directions",
    author = "Walhain, Lucy  and
      Albouze, S{\'e}bastien  and
      Gerencs{\'e}r, Anik{\'o}  and
      Paunescu, Mihai  and
      Tzouvaras, Vassilis  and
      Palma, Cosimo",
    editor = "Alam, Mehwish  and
      Tchechmedjiev, Andon  and
      Gracia, Jorge  and
      Gromann, Dagmar  and
      di Buono, Maria Pia  and
      Monti, Johanna  and
      Ionov, Maxim",
    booktitle = "Proceedings of the 5th Conference on Language, Data and Knowledge",
    month = sep,
    year = "2025",
    address = "Naples, Italy",
    publisher = "Unior Press",
    url = "https://aclanthology.org/2025.ldk-1.34/",
    pages = "340--350",
    ISBN = "978-88-6719-333-2"
}

@article{salatino2020cso,
  author = {Salatino, Angelo A. and Thanapalasingam, Thiviyan and Mannocci, Andrea and Birukou, Aliaksandr and Osborne, Francesco and Motta, Enrico},
  title = {The Computer Science Ontology: A Comprehensive Automatically-Generated Taxonomy of Research Areas},
  journal = {Data Intelligence},
  volume = {2},
  number = {3},
  pages = {379--416},
  year = {2020},
  doi = {https://doi.org/10.1162/dint_a_00055}
}

@misc{johnson2023mimic-note,
  title={MIMIC-IV-Note: Deidentified free-text clinical notes},
  author={Johnson, Alistair and Pollard, Tom and Horng, Steven and Celi, Leo Anthony and Mark, Roger},
  year={2023},
  publisher={PhysioNet}
}

@article{goldberger2000physiobank,
  title={PhysioBank, PhysioToolkit, and PhysioNet: components of a new research resource for complex physiologic signals},
  author={Goldberger, Ary L and Amaral, Luis AN and Glass, Leon and Hausdorff, Jeffrey M and Ivanov, Plamen Ch and Mark, Roger G and Mietus, Joseph E and Moody, George B and Peng, Chung-Kang and Stanley, H Eugene},
  journal={circulation},
  volume={101},
  number={23},
  pages={e215--e220},
  year={2000},
  publisher={Lippincott Williams \& Wilkins}
}

@article{Beltagy2020Longformer,
  title={Longformer: The Long-Document Transformer},
  author={Iz Beltagy and Matthew E. Peters and Arman Cohan},
  journal={arXiv:2004.05150},
  year={2020},
  url={https://arxiv.org/abs/2004.05150}
}

@inproceedings{xlm-roberta,
title = "Unsupervised Cross-lingual Representation Learning at Scale",
    author = "Conneau, Alexis  and
      Khandelwal, Kartikay  and
      Goyal, Naman  and
      Chaudhary, Vishrav  and
      Wenzek, Guillaume  and
      Guzm{\'a}n, Francisco  and
      Grave, Edouard  and
      Ott, Myle  and
      Zettlemoyer, Luke  and
      Stoyanov, Veselin",
    editor = "Jurafsky, Dan  and
      Chai, Joyce  and
      Schluter, Natalie  and
      Tetreault, Joel",
    booktitle = "Proceedings of the 58th Annual Meeting of the Association for Computational Linguistics",
    month = jul,
    year = "2020",
    address = "Online",
    publisher = "Association for Computational Linguistics",
    url = "https://aclanthology.org/2020.acl-main.747/",
    doi = "10.18653/v1/2020.acl-main.747",
    pages = "8440--8451",
}

@inproceedings{santosh2024chronoslex,
  title = "{C}hronos{L}ex: Time-aware Incremental Training for Temporal Generalization of Legal Classification Tasks",
    author = "T.y.s.s, Santosh  and
      Vuong, Tuan-Quang  and
      Grabmair, Matthias",
    editor = "Ku, Lun-Wei  and
      Martins, Andre  and
      Srikumar, Vivek",
    booktitle = "Proceedings of the 62nd Annual Meeting of the Association for Computational Linguistics (Volume 1: Long Papers)",
    month = aug,
    year = "2024",
    address = "Bangkok, Thailand",
    publisher = "Association for Computational Linguistics",
    url = "https://aclanthology.org/2024.acl-long.166/",
    doi = "10.18653/v1/2024.acl-long.166",
    pages = "3022--3039",
}

@article{agarwal2022temporal,
    title = "Temporal Effects on Pre-trained Models for Language Processing Tasks",
    author = "Agarwal, Oshin  and
      Nenkova, Ani",
    editor = "Roark, Brian  and
      Nenkova, Ani",
    journal = "Transactions of the Association for Computational Linguistics",
    volume = "10",
    year = "2022",
    address = "Cambridge, MA",
    publisher = "MIT Press",
    url = "https://aclanthology.org/2022.tacl-1.53/",
    doi = "10.1162/tacl_a_00497",
    pages = "904--921",
}

@inproceedings{niu2023towards_SAR,
  title={Towards Stable Test-time Adaptation in Dynamic Wild World},
  author={Niu, Shuaicheng and Wu, Jiaxiang and Zhang, Yifan and Wen, Zhiquan and Chen, Yaofo and Zhao, Peilin and Tan, Mingkui},
  booktitle={The Eleventh International Conference on Learning Representations},
  year={2023},
  url={https://openreview.net/pdf?id=g2YraF75Tj}
}

@inproceedings{niu2022EATA,
  title = 	 {Efficient Test-Time Model Adaptation without Forgetting},
  author =       {Niu, Shuaicheng and Wu, Jiaxiang and Zhang, Yifan and Chen, Yaofo and Zheng, Shijian and Zhao, Peilin and Tan, Mingkui},
  booktitle = 	 {Proceedings of the 39th International Conference on Machine Learning},
  pages = 	 {16888--16905},
  year = 	 {2022},
  editor = 	 {Chaudhuri, Kamalika and Jegelka, Stefanie and Song, Le and Szepesvari, Csaba and Niu, Gang and Sabato, Sivan},
  volume = 	 {162},
  series = 	 {Proceedings of Machine Learning Research},
  month = 	 {17--23 Jul},
  publisher =    {PMLR},
  url = 	 {https://proceedings.mlr.press/v162/niu22a.html},
}

@inproceedings{wang2021tent,
title={Tent: Fully Test-Time Adaptation by Entropy Minimization},
author={Dequan Wang and Evan Shelhamer and Shaoteng Liu and Bruno Olshausen and Trevor Darrell},
booktitle={International Conference on Learning Representations},
year={2021},
url={https://openreview.net/forum?id=uXl3bZLkr3c}
}

@inproceedings{huang2019neural,
  title = "Neural Temporality Adaptation for Document Classification: Diachronic Word Embeddings and Domain Adaptation Models",
    author = "Huang, Xiaolei  and
      Paul, Michael J.",
    editor = "Korhonen, Anna  and
      Traum, David  and
      M{\`a}rquez, Llu{\'i}s",
    booktitle = "Proceedings of the 57th Annual Meeting of the Association for Computational Linguistics",
    month = jul,
    year = "2019",
    address = "Florence, Italy",
    publisher = "Association for Computational Linguistics",
    url = "https://aclanthology.org/P19-1403/",
    doi = "10.18653/v1/P19-1403",
    pages = "4113--4123",
}

@inproceedings{shang2022improving,
  title = "Improving Time Sensitivity for Question Answering over Temporal Knowledge Graphs",
    author = "Shang, Chao  and
      Wang, Guangtao  and
      Qi, Peng  and
      Huang, Jing",
    editor = "Muresan, Smaranda  and
      Nakov, Preslav  and
      Villavicencio, Aline",
    booktitle = "Proceedings of the 60th Annual Meeting of the Association for Computational Linguistics (Volume 1: Long Papers)",
    month = may,
    year = "2022",
    address = "Dublin, Ireland",
    publisher = "Association for Computational Linguistics",
    url = "https://aclanthology.org/2022.acl-long.552/",
    doi = "10.18653/v1/2022.acl-long.552",
    pages = "8017--8026"
}

@inproceedings{shi2024medadapter,
    title = "{M}ed{A}dapter: Efficient Test-Time Adaptation of Large Language Models Towards Medical Reasoning",
    author = "Shi, Wenqi  and
      Xu, Ran  and
      Zhuang, Yuchen  and
      Yu, Yue  and
      Sun, Haotian  and
      Wu, Hang  and
      Yang, Carl  and
      Wang, May Dongmei",
    editor = "Al-Onaizan, Yaser  and
      Bansal, Mohit  and
      Chen, Yun-Nung",
    booktitle = "Proceedings of the 2024 Conference on Empirical Methods in Natural Language Processing",
    month = nov,
    year = "2024",
    address = "Miami, Florida, USA",
    publisher = "Association for Computational Linguistics",
    url = "https://aclanthology.org/2024.emnlp-main.1244/",
    doi = "10.18653/v1/2024.emnlp-main.1244",
    pages = "22294--22314"
}

@inproceedings{liu2025examining,
    title = "Examining and Adapting Time for Multilingual Classification via Mixture of Temporal Experts",
    author = "Liu, Weisi  and
      Han, Guangzeng  and
      Huang, Xiaolei",
    editor = "Chiruzzo, Luis  and
      Ritter, Alan  and
      Wang, Lu",
    booktitle = "Proceedings of the 2025 Conference of the Nations of the Americas Chapter of the Association for Computational Linguistics: Human Language Technologies (Volume 1: Long Papers)",
    month = apr,
    year = "2025",
    address = "Albuquerque, New Mexico",
    publisher = "Association for Computational Linguistics",
    url = "https://aclanthology.org/2025.naacl-long.313/",
    doi = "10.18653/v1/2025.naacl-long.313",
    pages = "6151--6166",
    ISBN = "979-8-89176-189-6"
}

@inproceedings{liu2025time,
  title={Time matters: Examine temporal effects on biomedical language models},
  author={Liu, Weisi and He, Zhe and Huang, Xiaolei},
  booktitle={AMIA Annual Symposium Proceedings},
  volume={2024},
  pages = {723--732},
  year={2024},
  pmid = {40417490},
  publisher = {American Medical Informatics Association},
  url={https://pmc.ncbi.nlm.nih.gov/articles/PMC12099427/},
  address = {San Francisco, CA, USA},
}

@inproceedings{huang2018modeling,
    title = "Modeling Temporality of Human Intentions by Domain Adaptation",
    author = "Huang, Xiaolei  and
      Liu, Lixing  and
      Carey, Kate  and
      Woolley, Joshua  and
      Scherer, Stefan  and
      Borsari, Brian",
    editor = "Riloff, Ellen  and
      Chiang, David  and
      Hockenmaier, Julia  and
      Tsujii, Jun{'}ichi",
    booktitle = "Proceedings of the 2018 Conference on Empirical Methods in Natural Language Processing",
    month = oct # "-" # nov,
    year = "2018",
    address = "Brussels, Belgium",
    publisher = "Association for Computational Linguistics",
    url = "https://aclanthology.org/D18-1074/",
    doi = "10.18653/v1/D18-1074",
    pages = "696--701",
}

@inproceedings{wolf2019huggingface,
  title = "Transformers: State-of-the-Art Natural Language Processing",
    author = "Wolf, Thomas  and
      Debut, Lysandre  and
      Sanh, Victor  and
      Chaumond, Julien  and
      Delangue, Clement  and
      Moi, Anthony  and
      Cistac, Pierric  and
      Rault, Tim  and
      Louf, Remi  and
      Funtowicz, Morgan  and
      Davison, Joe  and
      Shleifer, Sam  and
      von Platen, Patrick  and
      Ma, Clara  and
      Jernite, Yacine  and
      Plu, Julien  and
      Xu, Canwen  and
      Le Scao, Teven  and
      Gugger, Sylvain  and
      Drame, Mariama  and
      Lhoest, Quentin  and
      Rush, Alexander",
    editor = "Liu, Qun  and
      Schlangen, David",
    booktitle = "Proceedings of the 2020 Conference on Empirical Methods in Natural Language Processing: System Demonstrations",
    month = oct,
    year = "2020",
    address = "Online",
    publisher = "Association for Computational Linguistics",
    url = "https://aclanthology.org/2020.emnlp-demos.6/",
    doi = "10.18653/v1/2020.emnlp-demos.6",
    pages = "38--45",
}

@inproceedings{chalkidis2022improved,
    title = "Improved Multi-label Classification under Temporal Concept Drift: Rethinking Group-Robust Algorithms in a Label-Wise Setting",
    author = "Chalkidis, Ilias  and
      S{\o}gaard, Anders",
    editor = "Muresan, Smaranda  and
      Nakov, Preslav  and
      Villavicencio, Aline",
    booktitle = "Findings of the Association for Computational Linguistics: ACL 2022",
    month = may,
    year = "2022",
    address = "Dublin, Ireland",
    publisher = "Association for Computational Linguistics",
    url = "https://aclanthology.org/2022.findings-acl.192/",
    doi = "10.18653/v1/2022.findings-acl.192",
    pages = "2441--2454"
}

@article{dhingra2022time,
    title = "Time-Aware Language Models as Temporal Knowledge Bases",
    author = "Dhingra, Bhuwan  and
      Cole, Jeremy R.  and
      Eisenschlos, Julian Martin  and
      Gillick, Daniel  and
      Eisenstein, Jacob  and
      Cohen, William W.",
    editor = "Roark, Brian  and
      Nenkova, Ani",
    journal = "Transactions of the Association for Computational Linguistics",
    volume = "10",
    year = "2022",
    address = "Cambridge, MA",
    publisher = "MIT Press",
    url = "https://aclanthology.org/2022.tacl-1.15/",
    doi = "10.1162/tacl_a_00459",
    pages = "257--273"
}

@inproceedings{su2022improving,
    title = "Improving Temporal Generalization of Pre-trained Language Models with Lexical Semantic Change",
    author = "Su, Zhaochen  and
      Tang, Zecheng  and
      Guan, Xinyan  and
      Wu, Lijun  and
      Zhang, Min  and
      Li, Juntao",
    editor = "Goldberg, Yoav  and
      Kozareva, Zornitsa  and
      Zhang, Yue",
    booktitle = "Proceedings of the 2022 Conference on Empirical Methods in Natural Language Processing",
    month = dec,
    year = "2022",
    address = "Abu Dhabi, United Arab Emirates",
    publisher = "Association for Computational Linguistics",
    url = "https://aclanthology.org/2022.emnlp-main.428/",
    doi = "10.18653/v1/2022.emnlp-main.428",
    pages = "6380--6393"
}

@inproceedings{guo2023predict,
    title = "Predict the Future from the Past? On the Temporal Data Distribution Shift in Financial Sentiment Classifications",
    author = "Guo, Yue  and
      Hu, Chenxi  and
      Yang, Yi",
    editor = "Bouamor, Houda  and
      Pino, Juan  and
      Bali, Kalika",
    booktitle = "Proceedings of the 2023 Conference on Empirical Methods in Natural Language Processing",
    month = dec,
    year = "2023",
    address = "Singapore",
    publisher = "Association for Computational Linguistics",
    url = "https://aclanthology.org/2023.emnlp-main.65/",
    doi = "10.18653/v1/2023.emnlp-main.65",
    pages = "1029--1038"
}

@article{ji2026retrieval,
  title={Retrieval--Reasoning Processes for Multi-hop Question Answering: A Four-Axis Design Framework and Empirical Trends},
  author={Ji, Yuelyu and Li, Zhuochun and Meng, Rui and He, Daqing},
  journal={arXiv preprint arXiv:2601.00536},
  year={2026}
}

@article{ji2025mrag,
  title={MRAG-Suite: A Diagnostic Evaluation Platform for Visual Retrieval-Augmented Generation},
  author={Ji, Yuelyu and Lan, Wuwei and NG, Patrick},
  journal={arXiv preprint arXiv:2509.24253},
  year={2025}
}

@inproceedings{
wang2025reasoningretrieval,
title={Reasoning or Retrieval? A Study of Answer Attribution on Large Reasoning Models},
author={Yuhui Wang and Changjiang Li and Guangke Chen and Jiacheng Liang and Ting Wang},
booktitle={The Fourteenth International Conference on Learning Representations},
year={2026},
url={https://openreview.net/forum?id=DfxkLdy2Cd}
}

@inproceedings{margatina2023dynamic,
    title = "Dynamic Benchmarking of Masked Language Models on Temporal Concept Drift with Multiple Views",
    author = "Margatina, Katerina  and
      Wang, Shuai  and
      Vyas, Yogarshi  and
      Anna John, Neha  and
      Benajiba, Yassine  and
      Ballesteros, Miguel",
    editor = "Vlachos, Andreas  and
      Augenstein, Isabelle",
    booktitle = "Proceedings of the 17th Conference of the European Chapter of the Association for Computational Linguistics",
    month = may,
    year = "2023",
    address = "Dubrovnik, Croatia",
    publisher = "Association for Computational Linguistics",
    url = "https://aclanthology.org/2023.eacl-main.211/",
    doi = "10.18653/v1/2023.eacl-main.211",
    pages = "2881--2898"
}

@inproceedings{rottger2021temporal,
    title = "Temporal Adaptation of {BERT} and Performance on Downstream Document Classification: Insights from Social Media",
    author = {R{\"o}ttger, Paul  and
      Pierrehumbert, Janet},
    editor = "Moens, Marie-Francine  and
      Huang, Xuanjing  and
      Specia, Lucia  and
      Yih, Scott Wen-tau",
    booktitle = "Findings of the Association for Computational Linguistics: EMNLP 2021",
    month = nov,
    year = "2021",
    address = "Punta Cana, Dominican Republic",
    publisher = "Association for Computational Linguistics",
    url = "https://aclanthology.org/2021.findings-emnlp.206/",
    doi = "10.18653/v1/2021.findings-emnlp.206",
    pages = "2400--2412"
}

@inproceedings{Paszke2019pytorch,
 author = {Paszke, Adam and Gross, Sam and Massa, Francisco and Lerer, Adam and Bradbury, James and Chanan, Gregory and Killeen, Trevor and Lin, Zeming and Gimelshein, Natalia and Antiga, Luca and Desmaison, Alban and Kopf, Andreas and Yang, Edward and DeVito, Zachary and Raison, Martin and Tejani, Alykhan and Chilamkurthy, Sasank and Steiner, Benoit and Fang, Lu and Bai, Junjie and Chintala, Soumith},
 booktitle = {Advances in Neural Information Processing Systems},
 editor = {H. Wallach and H. Larochelle and A. Beygelzimer and F. d\textquotesingle Alch\'{e}-Buc and E. Fox and R. Garnett},
 pages = {},
 publisher = {Curran Associates, Inc.},
 title = {PyTorch: An Imperative Style, High-Performance Deep Learning Library},
 url = {https://proceedings.neurips.cc/paper_files/paper/2019/file/bdbca288fee7f92f2bfa9f7012727740-Paper.pdf},
 volume = {32},
 year = {2019}
}

@inproceedings{kutuzov2018diachronic,
    title = "Diachronic word embeddings and semantic shifts: a survey",
    author = "Kutuzov, Andrey  and
      {\O}vrelid, Lilja  and
      Szymanski, Terrence  and
      Velldal, Erik",
    editor = "Bender, Emily M.  and
      Derczynski, Leon  and
      Isabelle, Pierre",
    booktitle = "Proceedings of the 27th International Conference on Computational Linguistics",
    month = aug,
    year = "2018",
    address = "Santa Fe, New Mexico, USA",
    publisher = "Association for Computational Linguistics",
    url = "https://aclanthology.org/C18-1117/",
    pages = "1384--1397"
}

@inproceedings{tang-etal-2023-word,
    title = "Can Word Sense Distribution Detect Semantic Changes of Words?",
    author = "Tang, Xiaohang  and
      Zhou, Yi  and
      Aida, Taichi  and
      Sen, Procheta  and
      Bollegala, Danushka",
    editor = "Bouamor, Houda  and
      Pino, Juan  and
      Bali, Kalika",
    booktitle = "Findings of the Association for Computational Linguistics: EMNLP 2023",
    month = dec,
    year = "2023",
    address = "Singapore",
    publisher = "Association for Computational Linguistics",
    url = "https://aclanthology.org/2023.findings-emnlp.231/",
    doi = "10.18653/v1/2023.findings-emnlp.231",
    pages = "3575--3590"
}

@inproceedings{
jang2022towards,
title={Towards Continual Knowledge Learning of Language Models},
author={Joel Jang and Seonghyeon Ye and Sohee Yang and Joongbo Shin and Janghoon Han and Gyeonghun KIM and Stanley Jungkyu Choi and Minjoon Seo},
booktitle={International Conference on Learning Representations},
year={2022},
url={https://openreview.net/forum?id=vfsRB5MImo9}
}

@article{rao2026scoping,
	title = {A {Scoping} {Review} of {Synthetic} {Data} {Generation} by {Language} {Models} in {Biomedical} {Research} and {Application}: {Data} {Utility} and {Quality} {Perspectives}},
	issn = {2509-498X},
	url = {https://doi.org/10.1007/s41666-026-00229-9},
	doi = {10.1007/s41666-026-00229-9},
	journal = {Journal of Healthcare Informatics Research},
	author = {Rao, Hanshu and Liu, Weisi and Wang, Haohan and Huang, I-Chan and He, Zhe and Huang, Xiaolei},
	month = feb,
	year = {2026},
}

@inproceedings{han2025attributes,
    title = "Attributes as Textual Genes: Leveraging {LLM}s as Genetic Algorithm Simulators for Conditional Synthetic Data Generation",
    author = "Han, Guangzeng  and
      Liu, Weisi  and
      Huang, Xiaolei",
    editor = "Christodoulopoulos, Christos  and
      Chakraborty, Tanmoy  and
      Rose, Carolyn  and
      Peng, Violet",
    booktitle = "Findings of the Association for Computational Linguistics: EMNLP 2025",
    month = nov,
    year = "2025",
    address = "Suzhou, China",
    publisher = "Association for Computational Linguistics",
    url = "https://aclanthology.org/2025.findings-emnlp.1055/",
    doi = "10.18653/v1/2025.findings-emnlp.1055",
    pages = "19367--19389",
    ISBN = "979-8-89176-335-7"
}

\appendix

\section{Implementation Details}


\subsection{Temporal Data Partitioning}
\label{sec:temporal_partition}

The released MIMIC-IV-Notes data employs a de-identification mechanism that assigns the time information of each clinical note to three-year-long intervals spanning from 2008 to 2022, which naturally forms five temporal domains.
We selected the first four temporal intervals for our experiments due to data scarcity in the last time interval, denoted as $T_1, T_2, T_3, T_4$ in chronological order. 
To maintain consistency across datasets and enable systematic comparison, we adopt the same four-interval partitioning strategy. 
Table~\ref{tab:datasplit} presents the detailed temporal splits for each dataset.

The last interval is set as the target data, and earlier as the source data. 
Within each temporal interval, we randomly split the data into training (70\%) and test (30\%) sets. 
We train models on the training set of $T_1$ and evaluate on the test sets of both $T_1$ (in-domain) and $T_4$ (cross-domain) to test the temporal generalization of the model.

\begin{table}[htp]
\centering
\begin{tabular}{l|c}
\textbf{Dataset} & \textbf{Time Intervals} \\ \hline \hline
MIMIC    & 2008--2010, 2011--2016, 2017--2019 \\ \hline
EurLex   & 1958--1985, 1986--2010, 2011--2016 \\ \hline
arXiv-CS & 1991--2010, 2011--2020, 2021--2025 \\
\end{tabular}
\caption{Temporal splits of datasets.}
\label{tab:datasplit}
\end{table}

\subsection{Prompt Template }
\label{sec:prompt}
\lstset{
    backgroundcolor=\color[RGB]{245,245,245},
    breaklines=true,
    breakindent=0pt,
    basicstyle=\ttfamily\small,
    frame=single,           
    xleftmargin=6pt,        
    xrightmargin=6pt,
    framexleftmargin=6pt,
    framexrightmargin=6pt
}
\Needspace{0.5\textheight}
\begin{lstlisting}
You are helping to build a terminology lexicon for a multi-label
text classification task on the dataset: {dataset_name}.
You will receive:
- One document text.
- The full list of possible labels for this dataset.
Your tasks:
1. Read the document text carefully.
2. Choose 3-10 words or short phrases ("terms") that are especially
informative for deciding which labels to assign from the given label set.
These terms should:
- Be specific to the subject or content of the document, not generic words
like "data", "paper", "regulation", "method", "result", etc.
- Strongly indicate one or more labels when they appear.

2. For each selected term, generate:
- "synonyms": close paraphrases, especially earlier phrasings that could appear in older literature or legal/technical writing, preferred terms if applicable,
You MUST return a SINGLE JSON object with EXACTLY this structure:
{
"entities": [
{
"term": "...",
"synonyms": ["...", "..."]
}
]
}
No extra keys. No comments. No explanations outside JSON.
--------------------
DATASET NAME: {dataset_name}
FULL LABEL SET:
{label_block}
--------------------
DOCUMENT TEXT:
{text}
\end{lstlisting}

\subsection{External Knowledge Resources}
\label{sec:external_resources}

We leverage domain-specific ontology thesauri to provide structured
terminological relationships for data augmentation.

\textbf{MeSH (Medical Subject Headings)~\cite{mesh2025}}
We used the MeSH 2025 release \footnote{\url{https://www.nlm.nih.gov/mesh/}}, including both the descriptor file (\texttt{desc2025.xml}) and the supplementary concept file (\texttt{supp2025.xml}).
These XML files provide hierarchically-organized medical terminology and synonym relationships used for augmentation of the MIMIC-IV-Notes data.


\textbf{EuroVoc}~\cite{walhain-etal-2025-eurovoc}
We use version 4.22 of the EuroVoc thesaurus (released 2025-07-02), specifically the English version (\texttt{eurovoc\_export\_en-4.22.xlsx}). 
EuroVoc provides a multi-level hierarchical taxonomy of concepts spanning 21 domains and 127 sub-domains. 
The thesaurus defines relationships between preferred terms (PT), which represent canonical concept labels, and non-preferred terms (NPT), which serve as synonyms or alternative expressions. 
We exploit these PT-NPT relationships to identify terminological variants for data augmentation in EUR-Lex legal documents.


\textbf{CSO (Computer Science Ontology)~\cite{salatino2020cso}}
We use CSO version 3.5\footnote{\url{https://cso.kmi.open.ac.uk/}},
the latest release of the ontology.
We extract terminological relationships from CSO using the
CSO Classifier toolkit
(v4.0.0)~\cite{salatino2019classifier}\footnote{%
\url{https://github.com/angelosalatino/cso-classifier}},
which provides pre-processed ontology files including the
topic hierarchy and term-to-topic mappings.
These resources supplement the GPT-4o-mini-identified
terminology and generated synonyms for the arXiv-CS dataset.

\begin{algorithm}[t]
\caption{Knowledge-driven Augmentation and Retrieval for Integrative Temporal Adaptation}
\label{alg:karita}
\begin{algorithmic}[1]
\REQUIRE Source-trained model $\Theta_s$, source dataset $\mathcal{D}_s$, target dataset $\mathcal{D}_t$
\STATE Initialize model $\Theta \leftarrow \Theta_s$
\FOR{each target batch $\mathcal{B}_t \subset \mathcal{D}_t$}
    \STATE Detect shifted samples $\mathcal{B}_{shift} \subset \mathcal{B}_t$ using uncertainty, feature, and ontology-based scores
    \FOR{each $x_t \in \mathcal{B}_{shift}$}
        \STATE Compute embedding $\mathbf{z}_t \leftarrow f^{enc}_{\Theta}(x_t)$
        \STATE Retrieve top-$k$ source samples $\mathcal{R}(x_t)$ from $\mathcal{D}_s$ using cosine similarity
    \ENDFOR
    \STATE Augment retrieved samples using LLM-based and/or ontology-based synonym augmentation
    \STATE Update model $\Theta$ using the augmented source samples
\ENDFOR
\STATE \textbf{Prediction:} Generate predictions on $\mathcal{D}_t$ using the final adapted model $\Theta$
\end{algorithmic}
\end{algorithm}

\subsection{Experiment details and Baselines}

\paragraph{Source Model}
We initialize all methods from a common base model trained on the earliest time interval $T_1$. 
For EurLex and arXiv-CS, we use \textit{XLM-RoBERTa-base}~\cite{xlm-roberta} as the base encoder, while for MIMIC-IV we adopt Longformer~\cite{Beltagy2020Longformer} to better handle long clinical documents.
The source model is fine-tuned on gold-labeled data from $T_1$ for 10 epochs, using a learning rate of $3\times10^{-5}$. 
We use a batch size of 32 for EurLex and arXiv-CS, and 8 for MIMIC, with gradient accumulation steps set to 2.
Unless otherwise specified, we choose the same batch size and learning rate for all baseline methods and KARITA.

\paragraph{KARITA}
KARITA adapts the source-trained model sequentially over the target data stream.
For each target batch, shifted samples are first identified using the proposed shift detection module.
Only the detected samples trigger source backtracking retrieval, where semantically similar source instances are selected.
The retrieved samples are then augmented using knowledge-driven terminology expansion and used to update the model before processing the next target batch.
For efficiency, LLM-based terminology identification and synonym generation are performed once for the detected target samples and reused throughout adaptation, rather than being invoked within each update step.
After adaptation is completed, predictions are generated once on the full target split using the final adapted model.

\paragraph{Self-Labeling}
We follow the protocol in~\cite{agarwal2022temporal}. The source model fine-tuned on gold-labeled source data is first used to generate pseudo-labels for target-domain samples. 
The pseudo-labeled target data is then merged with the original source training set to fine-tune a new model. 

\paragraph{IFT}
For IFT~\cite{santosh2024chronoslex}, we train the model sequentially on chronologically ordered temporal splits.
We split the first interval of the data into three incremental splits and train on every split for 3 epochs.
The model parameters fine-tuned from the previous splits are used to initialize training for the next splits. 

\paragraph{SAR}~\cite{niu2023towards_SAR} We set the entropy minimization objective to operate on per-label sigmoid outputs to apply SAR to multi-label classification setting.  
Specifically, prediction entropy is computed independently for each label dimension and aggregated across labels. 
We adopt the sharpness-aware optimizer proposed in SAR to stabilize online updates and perform adaptation using unlabeled target samples at test time. 
For SAR, we use a learning rate of $1\times10^{-4}$ with a sharpness-aware optimizer, momentum set to $0.9$, and a perturbation radius of $\rho=0.05$. 
The entropy threshold is defined as $E_0 = 0.4 \cdot \ln(\texttt{max\_token\_len})$.

\paragraph{EATA}~\cite{niu2022EATA} Similar to SAR, we apply EATA to the multi-label setting by computing entropy over sigmoid-based label probabilities. 
Only target samples with low aggregated entropy are selected for model updating. 
In addition, we apply the regularization mechanism proposed in EATA to limit parameter drift and mitigate catastrophic forgetting during test-time adaptation. 
For EATA, we adopt a learning rate of $5\times10^{-5}$ and apply Fisher regularization with $\beta = 1/2000$ to constrain parameter updates. 
The entropy threshold is set to $E_0 = 0.4 \cdot \ln(\texttt{max\_token\_len})$ to select reliable target samples for adaptation.

\subsection{Hardware and Software}

The experiments  are conducted on a server equipped with 8x H100 GPUs, 2x EPYC Genoa 9334 CPUs, and 768GB of RAM. 
The system runs on Linux kernel 5.14.
The system utilizes PyTorch 2.3.0 (CUDA 12.1)~\cite{Paszke2019pytorch} alongside HuggingFace Transformers 4.57.1 ~\cite{wolf2019huggingface}. 

\section{Additional Results}

\subsection{Sensitivity Analysis}
\label{sec:sensitivity}

We examine the sensitivity of KARITA to two key
hyper-parameters: the number of retrieved source samples $k$
and the shift detection proportion $\rho$.
We conduct experiments on MIMIC-IV-Notes as a representative
case, varying one parameter while fixing the other.

\paragraph{Effect of $k$}
We vary $k$ from 1 to 5 with $\rho$ fixed at 0.1.
As shown in Table~\ref{tab:sensitivity_k}, performance
improves from $k{=}1$ to $k{=}3$ and remains relatively
stable beyond that, with $k{=}3$ achieving the best overall
balance across metrics.
Retrieving too few samples ($k{=}1$) limits the diversity
of source supervision, while larger values ($k{=}4, 5$) do
not yield further gains and increase computational cost.

\begin{table}[h]
\centering
\begin{tabular}{l|ccccc}
Metric & $k{=}1$ & $k{=}2$ & $k{=}3$ & $k{=}4$ & $k{=}5$ \\
\hline \hline
P     & 64.04 & 67.34 & 67.00 & 66.07 & 69.34 \\
R     & 55.74 & 61.18 & 63.35 & 63.55 & 60.87 \\
sa-F1 & 56.05 & 59.86 & 61.80 & 61.03 & 61.07 \\
mi-F1 & 59.60 & 64.11 & 65.13 & 64.78 & 64.83 \\
ma-F1 & 46.57 & 51.78 & 53.12 & 52.60 & 52.15 \\
\end{tabular}
\caption{Sensitivity analysis on $k$ ($\rho = 0.1$,
MIMIC-IV-Notes).}
\label{tab:sensitivity_k}
\end{table}

\paragraph{Effect of $\rho$}
We vary $\rho$ across $\{0.05, 0.1, 0.2, 0.3\}$ with $k$
fixed at 3.
As shown in Table~\ref{tab:sensitivity_rho}, performance
improves as $\rho$ increases from 0.05 to 0.1, but does not
improve further at 0.2 and 0.3.
A smaller $\rho$ limits the coverage of shifted samples,
while larger values introduce non-shifted samples into
retrieval without additional benefit.
We therefore choose $\rho{=}0.1$ as a practical balance
between effectiveness and efficiency.

\begin{table}[h]
\centering
\begin{tabular}{l|cccc}
Metric & $\rho{=}0.05$ & $\rho{=}0.1$ & $\rho{=}0.2$
       & $\rho{=}0.3$ \\ \hline \hline
P     & 68.14 & 67.00 & 65.43 & 67.75 \\
R     & 57.51 & 63.35 & 63.24 & 61.56 \\
sa-F1 & 58.76 & 61.80 & 61.10 & 61.41 \\
mi-F1 & 62.37 & 65.13 & 64.32 & 64.51 \\
ma-F1 & 49.05 & 53.12 & 53.68 & 53.33 \\
\end{tabular}
\caption{Sensitivity analysis on $\rho$ ($k = 3$,
MIMIC-IV-Notes).}
\label{tab:sensitivity_rho}
\end{table}



\end{document}